\documentclass[preprint]{elsarticle}

\usepackage{lineno,hyperref}
\usepackage[utf8]{inputenc}
\usepackage{amsfonts}
\usepackage{amsmath}
\modulolinenumbers[5]

\journal{Journal of Biomedical Informatics}

\begin{document}

\begin{frontmatter}

\title{ISeeU: Visually interpretable deep learning for mortality prediction inside the ICU}

\author[AUT,UTB]{William Caicedo-Torres\corref{mycorrespondingauthor}}
\cortext[mycorrespondingauthor]{Corresponding author}
\ead{william.caicedo@aut.ac.nz}

\author[AUT]{Jairo Gutierrez}

\address[AUT]{Auckland University of Technology, Auckland - New Zealand}
\address[UTB]{Universidad Tecnológica de Bolívar, Cartagena - Colombia}

\begin{abstract}
To improve the performance of Intensive Care Units (ICUs), the field of bio-statistics has developed scores which try to predict the likelihood of negative outcomes. These help evaluate the effectiveness of treatments and clinical practice, and also help to identify patients with unexpected outcomes. However, they have been shown by several studies to offer sub-optimal performance. Alternatively, Deep Learning offers state of the art capabilities in certain prediction tasks and research suggests deep neural networks are able to outperform traditional techniques. Nevertheless, a main impediment for the adoption of Deep Learning in healthcare is its reduced interpretability, for in this field it is crucial to gain insight on the why of predictions, to assure that models are actually learning relevant features instead of spurious correlations. To address this, we propose a deep multi-scale convolutional architecture trained on the Medical Information Mart for Intensive Care III (MIMIC-III) for mortality prediction, and the use of concepts from coalitional game theory to construct visual explanations aimed to show how important these inputs are deemed by the network. Our results show our model attains state of the art performance while remaining interpretable. Supporting code can be found at \url{https://github.com/williamcaicedo/ISeeU}.
\end{abstract}

\begin{keyword}
Deep Learning\sep MIMIC-III\sep ICU \sep Shapley Values
\end{keyword}

\end{frontmatter}

\section{Introduction}

Intensive Care Units (ICUs) have helped to make improvements in mortality, length of stay and complication rates among patients \cite{Johnson2016}, but they are costly to operate and sometimes skilled personnel to staff them seems to be in short supply \cite{Johnson2016}. For this reason, research efforts to better equip ICUs to handle patients in a more cost-effective manner are warranted.

The field of bio-statistics have produced throughout the years a series of predictive scores which try to quantify the likelihood of negative outcomes (i.e. death) in clinical settings. These tools are necessary to evaluate the effectiveness of treatments and clinical practice, and to identify patients with unexpected outcomes \cite{Rapsang:2014aa}. Scores as the Apache (in its several versions), SAPS, MODS and others have had moderate success \cite{Rapsang:2014aa}. Although their performance is not optimal, they have become de facto standards for severity and mortality risk prediction. These scores have been built using statistical techniques such as Logistic Regression, which are limited to the modeling of linear decision boundaries, when it is quite likely that the actual dynamics of the related biological systems does not respond to such prior. A reason for limiting the modeling to linear/additive techniques as Logistic Regression is that they tend to be readily interpretable, allowing medical staff to derive rules and gain insight over the reasons why such a score is predicting certain risk or mortality probability. However, such statistical approaches (APACHE, SAPS, MODS, etc.), have been shown by several studies to generalize sub-optimally \cite{Huang:2013aa, Paul:2013aa}. \cite{Paul:2013aa} show that over time, fixed scores’ performance tends to deteriorate (e.g. APACHE III-j over-predicts mortality in Australasia), and cite as possible reasons changes in medical practice and better care. It’s no wonder then, that ICU mortality prediction appears to have reached a plateau \cite{Johnson2016}.

On the other hand, Deep Learning offers state of the art capabilities in object recognition and several related areas, and those capabilities can be used to learn to detect patterns in patient data and predict the likelihood of negative outcomes. A reliable survival prediction system using Machine Learning concepts such as supervised fine-tuning (with pre-training that uses data from a related domain) and online learning (keep learning after deployment) could overcome the degradation problems exhibited by fixed scores, by being able to learn from the environments where they are being deployed. This would benefit ICUs everywhere, allowing staff to benchmark ICU performance and improve treatment protocols and practice \cite{Rapsang:2014aa}. 

Machine Learning models depend on data for training, and in the case of Deep Learning, the amount of data needed to reach adequate performance can be larger than what traditional Machine Learning models require. However, today there is a deluge of data coming from various disparate sources, and said data sometimes sits in databases without much use. In the case of Electronic Medical Records, detailed information about patients as visit records and socio-demographic data is stored indefinitely and could be leveraged to train predictive models that enable precision healthcare.

One of the main impediments for widespread adoption of advanced Machine Learning and Deep Learning in healthcare is lack of interpretability \cite{Che:2016aa, DBLP:journals/corr/Lipton16a}. There seems to be a trade-off between predictive accuracy and interpretability in the landscape of learning algorithms, and in the case of Deep Neural Networks, models of greater depth consistently outperform shallower ones in some tasks \cite{He:2016aa, Szegedy:2015aa, do_deep_convnets}, at the expense of simpler representations. Crucially, high capacity Machine Learning models can easily latch onto epistemically flawed correlations and statistical flukes as long as they help minimize the loss in the training set, because the minimization of the associated loss function does not care for causality but merely for correlation \cite{DBLP:journals/corr/Lipton16a}. For instance, in one well-known case a neural network \cite{Cooper1997} was trained to predict the risk of death in patients with pneumonia, and it was found that the model consistently predicted lower probability of death for patients who also had asthma. There was a counter-intuitive correlation in the training data that did not reflect any causality whatsoever, just the fact that asthma patients were treated more aggressively and thus fared better in average. The model in question performed better than the rest of models considered but it was ultimately discarded in favor of less performant, but interpretable ones. It is crucial then to offer mechanisms to gain insight on the why of predictions, i.e. the features our models attend to when generating an output, to make sure that models are actually learning sensible features instead of spurious and misleading correlations

In this paper, we propose a multi-scale deep convolutional architecture to tackle the problem of mortality prediction inside the ICU, trained on clinical data. One central feature of our approach is that it is geared to offer interpretable predictions, i.e. predictions accompanied by explanations and/or justifications which make for a more transparent decision process. For the latter we leverage the concept of Shapley Values \cite{shapley:book1952}, to create visualizations that convey the importance that the convolutional model assigns to each input feature. The relationship of this work with the existing literature and its main contributions are summarized next.

Our work relates to the existing literature in a number of ways. We use Deep Learning for mortality prediction inside the ICU as it also has been used by Che et al \cite{DBLP:journals/corr/ChePCSL16, Che:2016aa}, Grnarova et al \cite{Grnarova2016NeuralDE} and Purushotham et al \cite{PURUSHOTHAM2018}, but our work has key differences: 

\begin{itemize}
\item We are able to show that ConvNets offer predictive performance comparable to the reported performance of RNNs when dealing with physiological time-series data from MIMIC-III.
\item We show evidence that a deep convolutional architecture can handle both static and dynamic data from MIMIC-III, making hybrid architectures (DNN/RNN) unnecessary at this particular task and performance level. 
\item We achieve the previously mentioned results using simple forward/backward filling imputation and mean imputation instead of more involved and computationally expensive approaches.
\end{itemize}

Regarding the problem of interpretability, the work most related to ours is the one by Che et al \cite{Che:2016aa}. However, in this case there are also some important differences: 
\begin{itemize}
\item Che et al sidestep the problem of interpreting a deep model directly by using Mimic Learning with an interpretable student model (Gradient Boosted Trees) \cite{Ba:2014:DNR:2969033.2969123}, while our work focus instead on interpreting directly a deep model trained to predict ICU mortality, without using any surrogate model.  
\item We are able to provide not only dataset-level interpretability but also patient-level interpretability.
\item Our model works with raw features instead of pre-processed ones.
\end{itemize}

On the other hand, our architecture uses multi-scale convolutional layers and a "channel" input representation, similar to \cite{suresh2017clinical}, but for a different task (mortality prediction instead of clinical intervention prediction). We also note that the use of Shapley Values \cite{shapley:book1952} or their approximations for providing interpretability in the ICU setting has not, to the best our knowledge, been reported by the relevant literature.

\subsection{Related work} Although the most natural application of Deep Learning algorithms to medical diagnosis is automated medical image diagnosis \cite{annurev-bioeng-071516-044442}, the usage of Physiological Time Series (PTS) and Electronic Medical Record (EMR) data, is a more general source of data on which machine learning models can be trained. EMRs are very attractive as a potential data source since their use is widespread, which makes them abundant and accessible electronically. However, there are certain challenges associated with their “secondary use” in Machine Learning \cite{Johnson2016}. Despite this, several works have reported the successful use of EMRs and PTS to train Machine Learning/Deep Learning based models for diagnosis. 

In one of the first published attempts of using deep learning for medical tasks, Lasko et al \cite{Lasko:2013aa} used a deep Autoencoder for unsupervised clinical phenotype discovery from serum uric acid measurements. Che et al \cite{Che:2015:DCP:2783258.2783365} proposed a feed-forward deep model with sigmoidal activations to predict survival into the ICU, trained using data from the PhysioNet Challenge 2012 \cite{Silva:2012aa}. Kale et al \cite{Kale:2015aa} used Stacked Denoising Autoencoders (SDA) trained on the PhysioNet Challenge 2012 dataset and on a dataset extracted from the EMR system of the Children’s Hospital LA Pediatric Intensive Care Unit (PICU), to predict ICU mortality and diagnose 17 disease groups (according to ICU-9 hierarchical codes), respectively. Lipton et al \cite{Lipton:2015aa} used the Long Short-Term Memory Neural Network (LSTM) \cite{Hochreiter:1997:LSM:1246443.1246450} for multilabel classification of diagnoses inside Children’s Hospital Los Angeles PICU, in what according to the authors is the first work that uses LSTMs in a clinical prediction setting. In follow-up work \cite{pmlr-v56-Lipton16}, the authors treat the missing data problem from an interesting point of view. As similarly shown in \cite{RazavianS15}, the patterns of lab-testing in patients convey information in themselves. The authors explore the benefits of imputation against simple zero-filling, and modeling directly the “missingness” of data by concatenating a binary mask to the input at each time-step. Che et al \cite{DBLP:journals/corr/ChePCSL16} used a RNN based on a modified version of a Gated Recurrent Unit (GRU) to learn from multivariate time series with missing values (GRU-D). The authors used data from the MIMIC-III database \cite{Johnson:2016aa} and the PhysioNet Challenge 2012, as well as synthetic data to show the performance of their model. Grnarova et al \cite{Grnarova2016NeuralDE} proposed an interesting application of Natural Language Processing (NLP) to mortality prediction in ICU patients. Their approach consisted in the use of a ConvNet trained on free-text clinical notes extracted from the MIMIC-III database. Che et al \cite{Che:2016aa} propose the use of deep models as Deep feed-forward Networks (DNN) \cite{Rumelhart:1986aa} and the Gated Recurrent Unit (GRU) (Chung et al., 2014) to predict ICU outcomes. Their main contribution was the use of mimic learning \cite{Ba:2014:DNR:2969033.2969123} to distill the knowledge gained by the deep models into a shallow, explainable model. In \cite{DBLP:journals/corr/ChoiBSSS16a} Choi et al proposed the use of medical ontologies (i.e. ICD-9) formulated as Directed Acyclical Graphs (DAGs) to regularize deep learning models via an attention mechanism. In \cite{suresh2017clinical}, Suresh et al use deep networks that leverage demographic information, physiological time series data and pre-computed representations of clinical notes extracted from MIMIC-III to predict the onset and weaning of medical interventions (invasive ventilation, non-invasive ventilation, vasopressors, colloid boluses, and crystalloid boluses) inside the ICU. Aczon et al \cite{Aczon2017} trained a recurrent neural network to generate temporally dynamic ICU mortality predictions at user-specified times. Beaulieu-Jones et al \cite{Beaulieu-Jones2018} used MIMIC-III data to train different deep learning models to perform one-year survival prediction. Finally, Purushotham et al \cite{PURUSHOTHAM2018} carried out a comprehensive benchmark of several machine learning and deep learning models trained on MIMIC-III for various tasks, with results showing deep models consistently outperforming the rest.

\section{Methods and materials}
\subsection{Participants} We used the Medical Information Mart for Intensive Care III (MIMIC-III v1.4) to train our deep models. MIMIC III is a database comprised of more than a decade worth of different modalities of detailed data from patients admitted to the Beth Israel Deaconess Medical Centre in Boston, Massachusets, freely available for research \cite{Johnson:2016aa}. To establish a cohort and build our dataset, several entry criteria were established: we only considered stays longer than 48 hours, only patients older than 16 years old at the time of admission were included, and in case of multiple admissions to the ICU, only the first one was considered. Application of these entry criteria led to a dataset containing 22.413 distinct patients.
\begin{table}[h]

\centering

\resizebox{\columnwidth}{!}{%
\begin{tabular}{ l r r r r r r r r r}
  \hline			
   Feature & n & Mean & Std. & Min & Q1 & Q2 & Q3 & Max & Percent of total pop. \\
   \hline	
   \textbf{Temporal} & &\\
   \hline
   Bicarbonate	& 99668	& 23.296 & 4.733 & 5.000 	& 20.000 & 23.000 & 26.000 & 52.000 & N/A\\
   Bilirrubin & 24765 	& 3.098	& 6.170 & 0.100	& 0.500 & 1.000	& 2.900 & 82.000 & N/A\\
   Bun & 101133  & 27.497 & 22.493 & 1.000	& 13.000	 & 20.000 & 34.000 & 240.000 & N/A\\
   Diastolic BP & 1752075 & 59.648 & 14.090 & 1.000 & 50.000 & 58.000 & 68.000 & 298.000 & N/A\\
   FiO2 & 294740 & 50.100 & 20.030 & 0.400 & 40.000 & 50.000 & 50.000 & 100.000 & N/A\\
   GCSEyes & 442646 & 3.140 &  1.142 &	1.000 & 3.000 &	4.000 & 4.000 &	4.000 & N/A\\
   GCSMotor	& 440486 & 5.256 &	1.440 & 1.000 &	5.000 & 6.000 &	6.000 & 6.000 & N/A\\
   GCSVerbal	 & 441269 & 3.123 &	1.902 & 1.000 & 1.000 & 4.000 & 5.000 & 5.000 & N/A\\
   Heart rate & 1755492 &	87.913 &	18.951 &  0.350 & 75.000 &	86.000 &	99.000 &	280.000 & N/A\\
   PO2 & 160600 & 149.763 & 95.724 &	14.000 &	87.000 &	119.000 &177.000 &	763.000 & N/A\\
   Potassium &	176528 & 4.192 &	0.700 & 0.600 &	3.700 & 4.100 &  4.500 &	26.500 & N/A\\
   Sodium & 125440 & 138.294 &	5.350 & 1.210 &	135.000 & 138.000 &141.000 & 183.000 & N/A\\
   Systolic BP & 1755083 &	118.518 & 22.973 & 0.150 & 102.000 & 116.000 & 133.000 & 323.000 & N/A\\
   Temperature & 563035 & 37.007 & 0.8610 & 15.000 & 36.444 & 37.000 & 37.600 & 42.222 & N/A\\
   Urine output & 947826 & 113.900 & 162.357 & -4000.000 &  37.000 & 70.000 & 140.000 &	4800.000 & N/A\\
   WBC & 94209 & 12.743 & 11.377 & 0.100 & 8.000 &	11.200 &	15.200 &	528.000 & N/A\\
   	\hline
   \textbf{Static} & & \\
   \hline
   Age & 22413 & 63.828 & 15.576 & 16.016 & 53.998 & 67.1016 & 78.533 & 80.000 & N/A\\
   Elective admission & 3618 & N/A & N/A & N/A & N/A & N/A & N/A & N/A & 14.134\%\\
   Surgical admission & 8030 & N/A & N/A & N/A & N/A & N/A & N/A & N/A & 35.827\%\\
   AIDS & 113 & N/A & N/A & N/A & N/A & N/A & N/A & N/A & 0.504\%\\
   Metastatic cancer & 688 & N/A & N/A & N/A & N/A & N/A & N/A & N/A & 3.069\%\\
   Lymphoma & 317 & N/A & N/A & N/A & N/A & N/A & N/A & N/A & 1.414\%\\
   Mortality & 2185 & N/A & N/A & N/A & N/A & N/A & N/A & N/A & 9.748\%\\

  \hline  
\end{tabular}%

}
\caption{\label{dataset}Some dataset statistics.}
\end{table}

\begin{figure}[h]
  
  \centering
    \includegraphics[width=\textwidth]{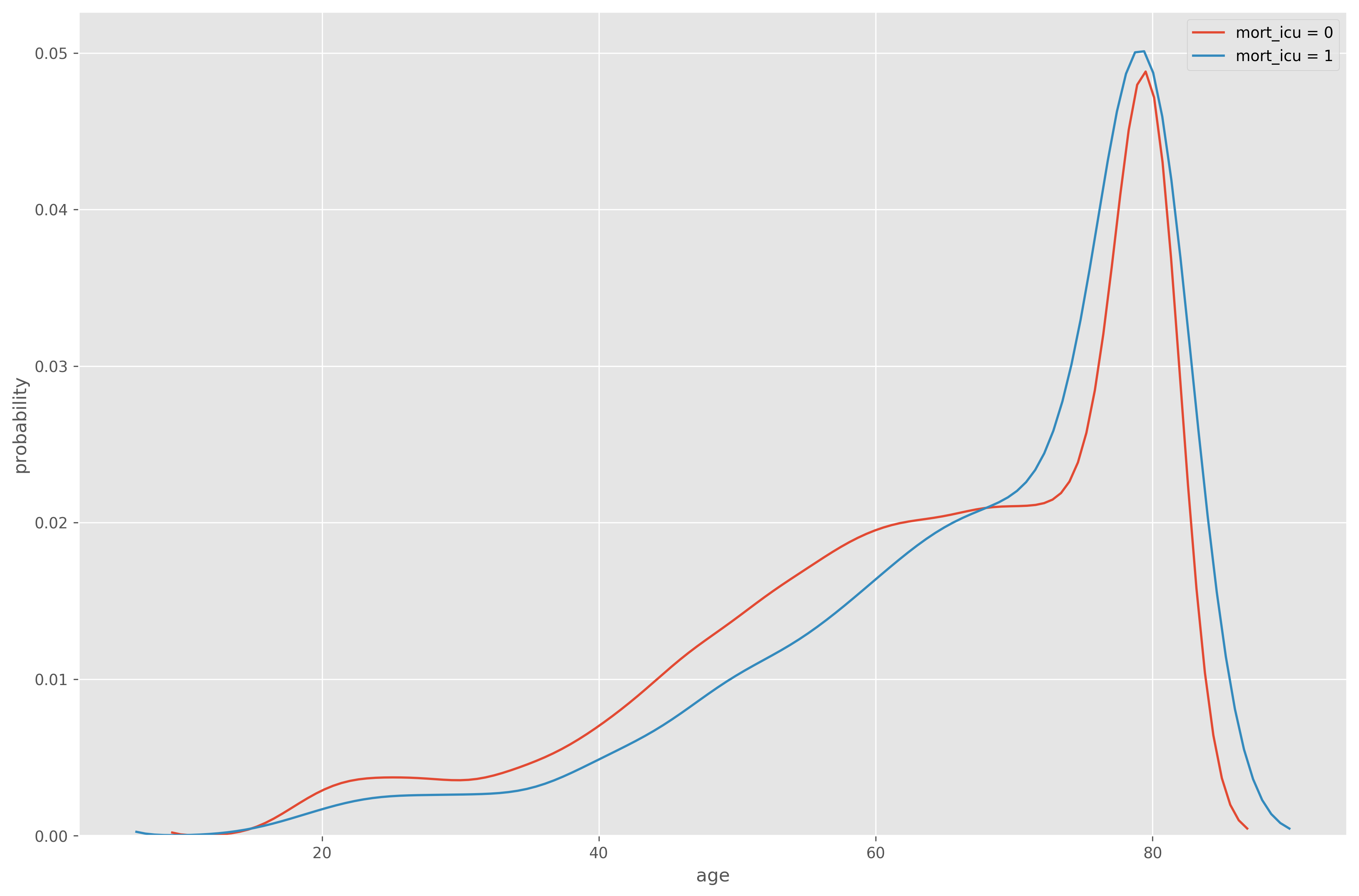}
    \caption{\label{kde-age-dist} KDE-generated age distribution, grouped by mortality. The large bump close to 80 years of age reflects our pre-processing.  }
\end{figure}

\begin{figure}[h]
  
  \centering
    \includegraphics[width=\textwidth]{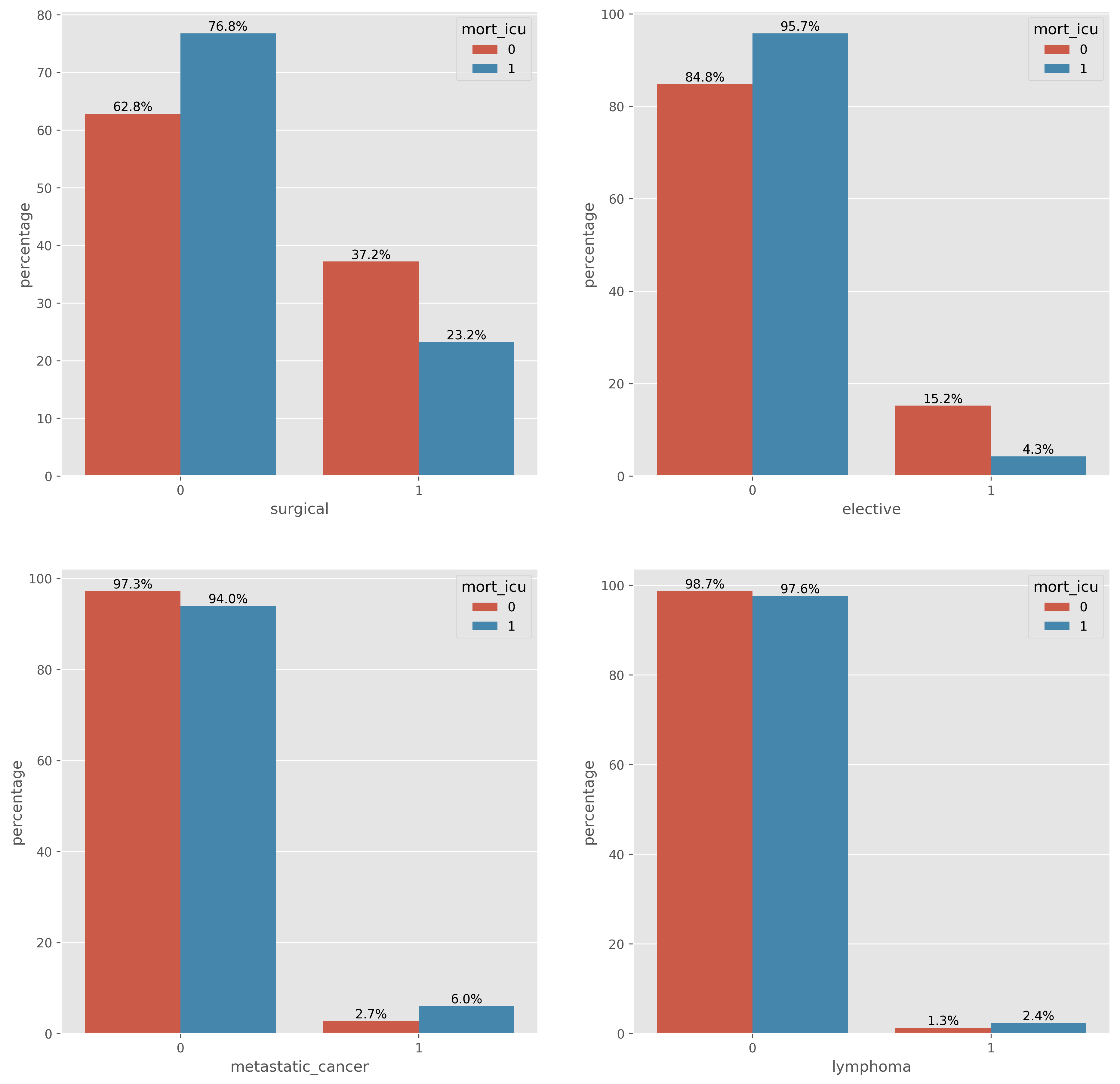}
    \caption{\label{feature-stats} Feature stats grouped by mortality }
\end{figure}

\subsection{Input features}
For each patient, we extracted measurements of 22 different concepts roughly matching the concepts used by the SAPS-II score \cite{Gall1993}, during the first 48 hours of each patient stay. In the case of temporal data, all measurements were extracted and in case of multiple measurement in the same hour, values were averaged (except urine output which was summed). To resolve inconsistencies and harmonize sometimes seemingly disparate concepts (i.e. temperature is reported both in Celsius and Fahrenheit, different codes for the same measurement are used, same or related concepts are present in different tables, etc), data was pre-processed, measurements merged and renamed, and un-physiological values were discarded. For privacy reasons, MIMIC-III shifts ages greater than 89 years (i.e. patients appear to be 300 years old). To address this, we clipped all ages greater than 80 years to 80 years (see figure \ref{kde-age-dist}). For reproducibility, all of our code is available at \url{https://github.com/williamcaicedo/MortalityPrediction}.

\begin{table}[hbt!]
\centering
\begin{tabular}{ l l }
  \hline			
   Feature & MIMIC-III Table \\
   \hline		
  Age  & ICUSTAYS, PATIENTS \\
  Presence of AIDS & DIAGNOSES\_ICD\\
  Blood bicarbonate & LABEVENTS\\
  Blood bilirubin & LABEVENTS\\
  Blood Urea Nitrogen & LABEVENTS\\
  Diastolic blood pressure & CHARTEEVENTS\\
  Systolic blood pressure & CHARTEEVENTS\\
  Admission to the ICU after surgery & SERVICES \\
  Elective admission to the ICU & ADMISSIONS\\
  Fraction of inspired oxygen (FIO2) & CHARTEVENTS, LABEVENTS\\
  Glasgow comma scale (eyes) & CHARTEVENTS\\
  Glasgow comma scale (motor) & CHARTEVENTS\\
  Glasgow comma scale (verbal) & CHARTEVENTS\\
  Heart rate & CHARTEVENTS\\
  Presence of lymphoma & DIAGNOSES\_ICD\\
  Presence of metastatic cancer & DIAGNOSES\_ICD \\
  Oxygen pressure in blood (PO2) & LABEVENTS\\
  Blood potassium & LABEVENTS\\
  Blood sodium & LABEVENTS\\
  Temperature & LABEVENTS\\
  Urine output & OUTPUTEVENTS\\
  White cell count & LABEVENTS\\
  \hline  
\end{tabular}
\caption{\label{dataset-features} Features extracted from MIMIC-III for each patient.}
\end{table}

\paragraph{Missing data} Due to the nature of patient monitoring, different physiological variables and features are sampled at different rates. This lead to large number of missing observations, as not all measurements were available hourly. Given this situation, simple data imputation techniques were applied to obtain a $22x48$ observation matrix for each patient (static features like age and admission where replicated). Concretely, except for urine observations, forward/backward filling imputation was attempted. After this, outstanding missing FiO2 values were then imputed to their normal values. On the other hand, when multiple observations are present in the same hour, values were averaged. In cases where a patient did not have a single observation recorded, we imputed the whole physiological time series using the empirical mean. Our data imputation procedure is summarized in table \ref{imputation}.

\begin{table}[hbt!]

\centering

\resizebox{\columnwidth}{!}{%
\begin{tabular}{l r p{10cm}}
  \hline			
   Feature & \% of missing values & Imputation procedure \\
   \hline	
   Bicarbonate	& 92.79\%	& Forward/Backward filling, mean value imputation\\
   Bilirrubin & 98.23\% & Forward/Backward filling, mean value imputation\\
   BUN & 92.69\%  & Forward/Backward filling, mean value imputation\\
   Diastolic BP & 10.07\% & Forward/Backward filling, mean value imputation\\
   FiO2 & 83.04\% & Forward filling, normal value (0.2) imputation\\
   GCSEyes & 68.30\% & Forward filling imputation, mean value imputation\\
   GCSMotor	& 68.45\% & Forward filling imputation, mean value imputation\\
   GCSVerbal	 & 68.39\% & Forward filling imputation, mean value imputation\\
   Heart rate & 7.53\% &	Forward/Backward filling, mean value imputation\\
   PO2 & 89.35\% & Forward/Backward filling, mean value imputation\\
   Potassium &	88.21\% & Forward/Backward filling, mean value imputation\\
   Sodium & 91.27\% & Forward/Backward filling, mean value imputation\\
   Systolic BP & 10.05\% &	Forward/Backward filling, mean value imputation\\
   Temperature & 66.15\% & Forward/Backward filling, mean value imputation\\
   Urine output & 33.15\% & Mean value imputation\\
   WBC & 93.27\% & Forward/Backward filling, mean value imputation\\
  
  \hline  
\end{tabular}%

}
\caption{\label{imputation}Imputation procedure summary.}
\end{table}

\subsection{Deep Learning model}
Our prediction model is a multi-scale deep convolutional neural network (ConvNet). ConvNets are Multi-Layer Neural Networks that use a particular architecture with sparse connections and parameter sharing \cite{LeCun:1998aa}. They can be thought of performing a discrete convolution operation between the input (often a two-dimensional image) and a set of trainable kernels at each layer. The discrete convolution operation, in the context of Deep Learning and computer vision is defined as

\begin{equation}
	s(i,j)=(x*w)(i,j)=\sum_{m,n}I(m,n)K(i-m,j-n)
\end{equation}

where I is a two-dimensional image and K is a two-dimensional kernel. The kernel acts as a local feature detector that is displaced all over the image. Each convolution between the input and a kernel produces a spatial receptive field, also called a feature map, in which each kernel-image multiplication can be thought of as pattern matching, which produces an output that is a function of the similarity between certain image region and the kernel itself. After the convolution operation, the output of the receptive field is ran through a non-linear activation function which allows the network to work with transforms of the input space and construct non-linear features. The feature map can be thought as a 2-D tensor (matrix) of neuron outputs, where the weights of each neuron are the same but have been shifted spatially (hence the parameter sharing), and which are not connected to every single pixel of the input (which also can be seen as having the corresponding weight set to zero). ConvNets were one the first models to use Gradient Descent with Backpropagation \cite{Rumelhart1986} with success \cite{Goodfellow-et-al-2016}. Convolution based filters are extensively used to detect features as shapes and edges in computer vision \cite{shapiro2001computer}. However, in traditional computer vision fixed kernels are used to detect specific features, in contrast to ConvNets where kernels are learned directly from the data. \\

\begin{figure}[h]
  
  \centering
    \includegraphics[width=\textwidth]{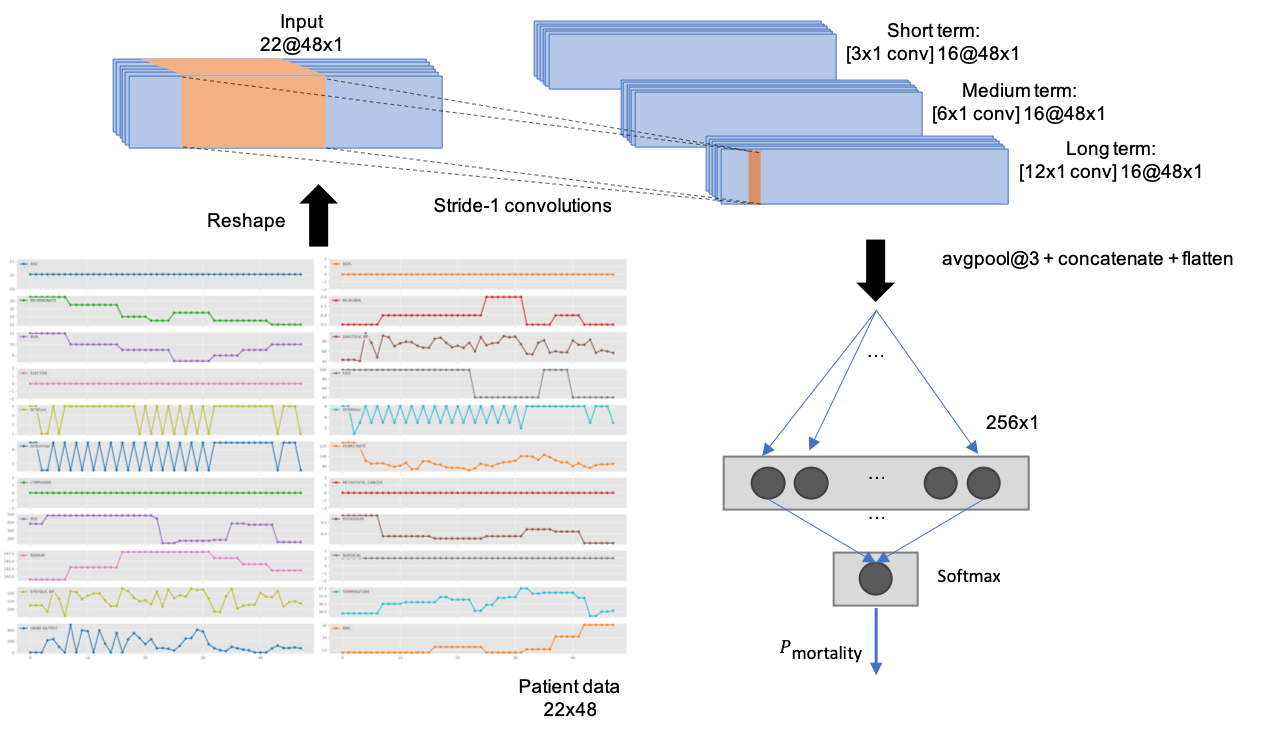}
    \caption{\label{architecture} Convolutional architecture for mortality prediction. ReLU activations, BatchNorm, and Dropout layers, have been omitted for clarity and brevity. }
\end{figure}
We define our model as a multi-scale ConvNet, as we use convolution kernels with different sizes and then concatenate the resulting feature maps into a single layer output tensor (figure \ref{architecture}). To deal with the different characteristics of our input time series, we employ a multi-scale convolutional layer, followed by ReLU activations, average-pooling with a window size of 3 and dropout \cite{Srivastava:2014} plus Batch Normalization \cite{IoffeS15} performed after the concatenation operation. In this layer we employ three temporal scales: Three hours, six hours, and twelve hours; each represented by a stack of convolution kernels with dimensions 3x1, 6x1, and 12x1, respectively. The convolutional layer is followed by a fully connected layer with ReLU activations, Dropout, Batch Normalization and a final one-neuron layer with logistic activation. Finally, our input representation places each feature as an image channel instead of stacking them as a 2-D input. This allows us to use 1-D temporal convolutions no matter how many input series we use.

\subsection{Shapley Values and input relevance attribution} The Shapley Value \cite{shapley:book1952} is a concept from game theory that formalizes the individual contribution of a coalition of players to the attainment of a reward in a game \cite{Strumbelj2010}. Shapley Values are the expectation of such contribution over the set of all possible permutations and values of the player coalition, taking into consideration all possible interactions between players. Formally, for a coalitional form game $\langle N, v\rangle$, where $N$ is a finite set of players and $ v:2^N \rightarrow \mathbb{R}$ describes the worth of a player coalition, we have that

\begin{equation} \label{shapley}
Sh_i(v)=\sum_{S \subseteq N\setminus\{i\},s= |S|}\frac{(n-s-1)!s!}{n!}(v(S \cup \{i\})-v(S))
\end{equation}

where $Sh_i$ is the individual contribution of player $i$ to the total worth $v(N)$, i.e. its Shapley value \cite{shapley:book1952}. The summation runs over all possible subsets of players $S \subseteq N$ that don't include player $i$, and each term involves the difference between the reward when player $i$ is present and absent, $v(S \cup \{i\})-v(S)$. Equation \ref{shapley} not only considers the presence of a particular player, but also the position it occupies in the coalition. This is extremely useful in the context of our study, where values are time sensitive.

Strumbelj et al \cite{Strumbelj2010} shown that such values can be used to represent the relevance of each input to a machine learning classifier in order to gain insight on the patterns it considers important to predict a particular class and proposed a feature importance attribution method equivalent to calculating the Shapley values. It is worth mentioning that the use of Shapley values for importance attribution is able to take into account the possible interactions between input features in a way occlusion-based methods \cite{Zeiler:2014aa} cannot.

\subsubsection{DeepLIFT} Computing equation \ref{shapley} has combinatorial cost, making it unfeasible for several practical applications, reason why we must resort to approximations. In this context, we will discuss a new importance attribution method, called DeepLIFT. DeepLIFT \cite{DBLP:journals/corr/ShrikumarGK17} is an importance attribution method for feed forward neural networks, that is akin to the Layer-wise Relevance Propagation method (LRP) proposed by \cite{Bach2015}, in the sense that both use a backpropagation-like approach to the calculation and attribution of relevance/importance scores to the input features. DeepLIFT overcomes problems associated with gradient-based attribution methods \cite{SimonyanVZ13, DBLP:journals/corr/SpringenbergDBR14} as saturation, overlooking negative contributions and contributions when the associated gradient is zero, and discontinuities in the gradients \cite{DBLP:journals/corr/ShrikumarGK17}. Since the attribution output of LRP was later shown to be roughly equivalent to a factor of a gradient method’s output \cite{2016arXiv161107270K}, it follows that LRP suffers from similar problems to those outlined before. 

\begin{figure}[h]
  
  \centering
    \includegraphics[scale=0.4]{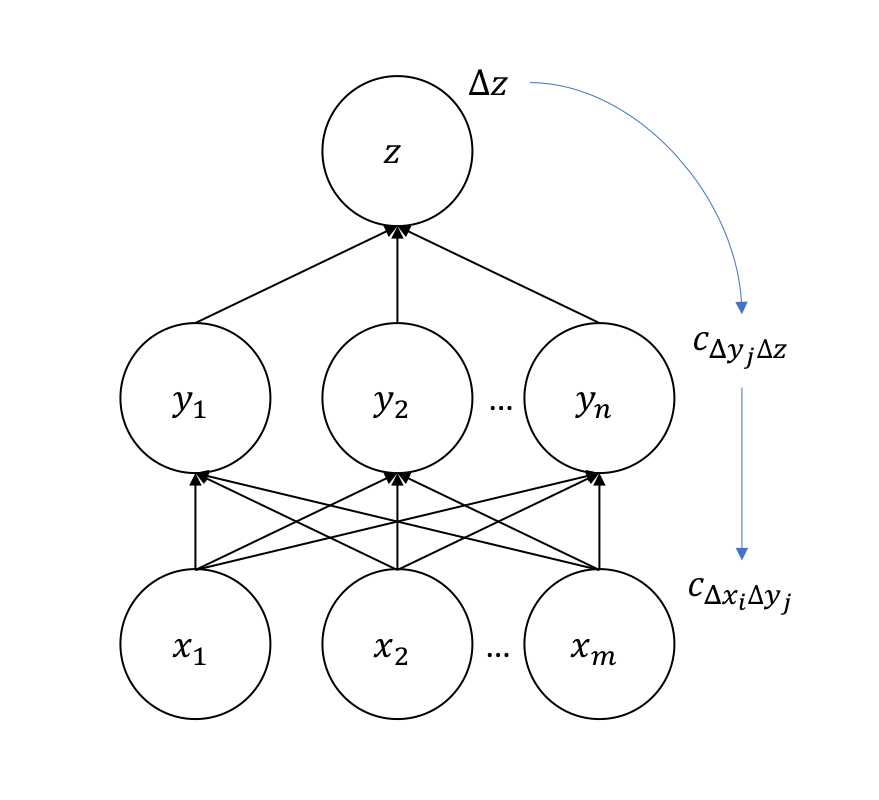}
    \caption{\label{deeplift} DeepLIFT's multipliers and chain rule allows the propagation of feature importances. }
\end{figure}

To compute feature importance the following procedure is carried out: first a reference input value must be provided. This reference value can be informed by domain knowledge or simply be the empirical mean of the input features, and once the references have been defined the corresponding network output is computed for both the original input and the reference input. Then the difference between outputs is calculated and backpropagated through the network layers using rules provided by DeepLIFT. This results in importance values that capture how a change in inputs contribute to the observed change in the output.

More formally, for a target neuron $t$ and a collection of neurons $x_1, x_2, ..., x_n$ whose outputs are needed to compute the output of $t$, the method assigns importance attributions $C_{\Delta x_i \Delta t}$ subject to the fact that such attributions are additive and must satisfy 
\begin{equation} \label{deltasum}
\sum_{i=1}^{n}C_{\Delta x_i \Delta t} = \Delta t
\end{equation}
where  $\Delta t = t_o - t_r$ is the difference between the original and reference outputs of $t$. DeepLIFT introduces multipliers $m_{\Delta x_i \Delta t} = \frac{c_{\Delta x_i \Delta t}}{\Delta x}$ that allow to use a chain-rule to backpropagate the neuron attributions through a hidden layer. The rule takes the form

\begin{equation}\label{deepchain}
m_{\Delta x_i \Delta z} = \sum_{j} m_{\Delta x_i \Delta y_j}m_{\Delta y_j \Delta z}
\end{equation}

where $m_{\Delta x_i \Delta z}$ is the contribution of neuron $x_i$ to the output of neuron $z$ divided by the difference in outputs for neuron $x_i$, $\Delta x_i$, given a hidden layer of neurons $y_j$ in-between (see figure \ref{deeplift}). The corresponding contribution $c_{\Delta x_i \Delta z}$ can be recovered from equation \ref{deepchain} as $c_{\Delta x_i \Delta z} = m_{\Delta x_i \Delta z}\Delta x$.

For a linear unit, the contribution of the inputs $x_i$ to the output difference $\Delta y $ is simply =  $ w_{i}\Delta x_i$. To avoid the issues that affect other methods regarding negative contributions, DeepLIFT treats separately positive and negative contributions, which leads to $\Delta y $ and $x_i$ being decomposed into its positive and negative components

\begin{equation}
\Delta y^+ = \sum_{i} 1\{w_i \Delta x_i > 0\}w_i (\Delta x_{i}^+ + \Delta x_{i}^-)
\end{equation}
\begin{equation}
\Delta y^- = \sum_{i} 1\{w_i \Delta x_i < 0\}w_i (\Delta x_{i}^+ + \Delta x_{i}^-)
\end{equation}

The contributions can be stated then as

\begin{equation}
c_{\Delta x_{i}^+ \Delta y^+ }= \sum_{i} 1\{w_i \Delta x_i > 0\}w_i \Delta x_{i}^+
\end{equation}
\begin{equation}
c_{\Delta x_{i}^- \Delta y^+ }= \sum_{i} 1\{w_i \Delta x_i > 0\}w_i \Delta x_{i}^-
\end{equation}
\begin{equation}
c_{\Delta x_{i}^+ \Delta y^- }= \sum_{i} 1\{w_i \Delta x_i < 0\}w_i \Delta x_{i}^+
\end{equation}
\begin{equation}
c_{\Delta x_{i}^- \Delta y^- }= \sum_{i} 1\{w_i \Delta x_i < 0\}w_i \Delta x_{i}^-
\end{equation}

For non-linear operations with a single input (e.g. ReLU activations), DeepLIFT proposes the so-called RevealCancel rule, which is able to better uncover non-linear dynamics \cite{DBLP:journals/corr/ShrikumarGK17}. For this case, $\Delta y$ decomposes as

\begin{equation}
\Delta y^+ = \frac{1}{2}(f(x_0 + \Delta x^+) - f(x_0)) + \frac{1}{2}(f(x_0 + \Delta x^- + \Delta x^+) - f(x_0 + \Delta x^-))
\end{equation}
\begin{equation}
\Delta y^- = \frac{1}{2}(f(x_0 + \Delta x^-) - f(x_0)) + \frac{1}{2}(f(x_0 + \Delta x^+ + \Delta x^-) - f(x_0 + \Delta x^+))
\end{equation}

And to satisfy \ref{deltasum} we have that $\Delta y^+ = c_{\Delta x_{i}^+ y^+}$ and $\Delta y^- = c_{\Delta x_{i}^- y^-}$. Given this, the multipliers for the RevealCancel rule are 
\begin{equation}
m_{\Delta x^+ y^+} = \frac{c_{\Delta x_{i}^+ y^+}}{\Delta y^+} = \frac{\Delta y^+}{\Delta y^+}
\end{equation}

\begin{equation}
m_{\Delta x^- y^-} = \frac{c_{\Delta x_{i}^- y^-}}{\Delta x^-}= \frac{\Delta y^-}{\Delta y^-}
\end{equation}

What makes DeepLIFT especially relevant is that it has been shown by Lundberg et al \cite{NIPS2017_7062} that DeepLIFT can be understood as fast approximation to the real Shapley Values when the feature reference values are set to their expected values. It can be seen that the RevealCancel rule computes the Shapley Values of the positive and negative contributions at the non-linear operations, and the successive application of the chain rule proposed by DeepLIFT allows to propagate the approximate Shapley Values back to the inputs.

\section{Results}

We built our ConvNet using Keras \cite{Chollet2015a} with Tensorflow \cite{USENIXAssociation.2015} as back-end. Since our dataset is highly unbalanced with the positive class (death) representing just under 10\% of training examples, we used a weighted logarithmic loss giving more weight to positive examples (1:10 importance ratio). We used 5-fold cross validation for a more reliable performance estimate and we standardized the dataset ($\mu = 0, \sigma \approx 1$) calculating fold statistics independently to avoid data leakage. We did not perform any substantial hyper-parameter optimization and instead opted for heuristically chosen values (dropout probability $0.45$, and a batch size of 32). Our choice of optimizer Stochastic Gradient Descent with Nesterov Momentum $0.9$, and learning rate of $0.01$ with a $1e-7$ decay.

\begin{figure}[h]
  
  \centering
    \includegraphics[scale=0.2]{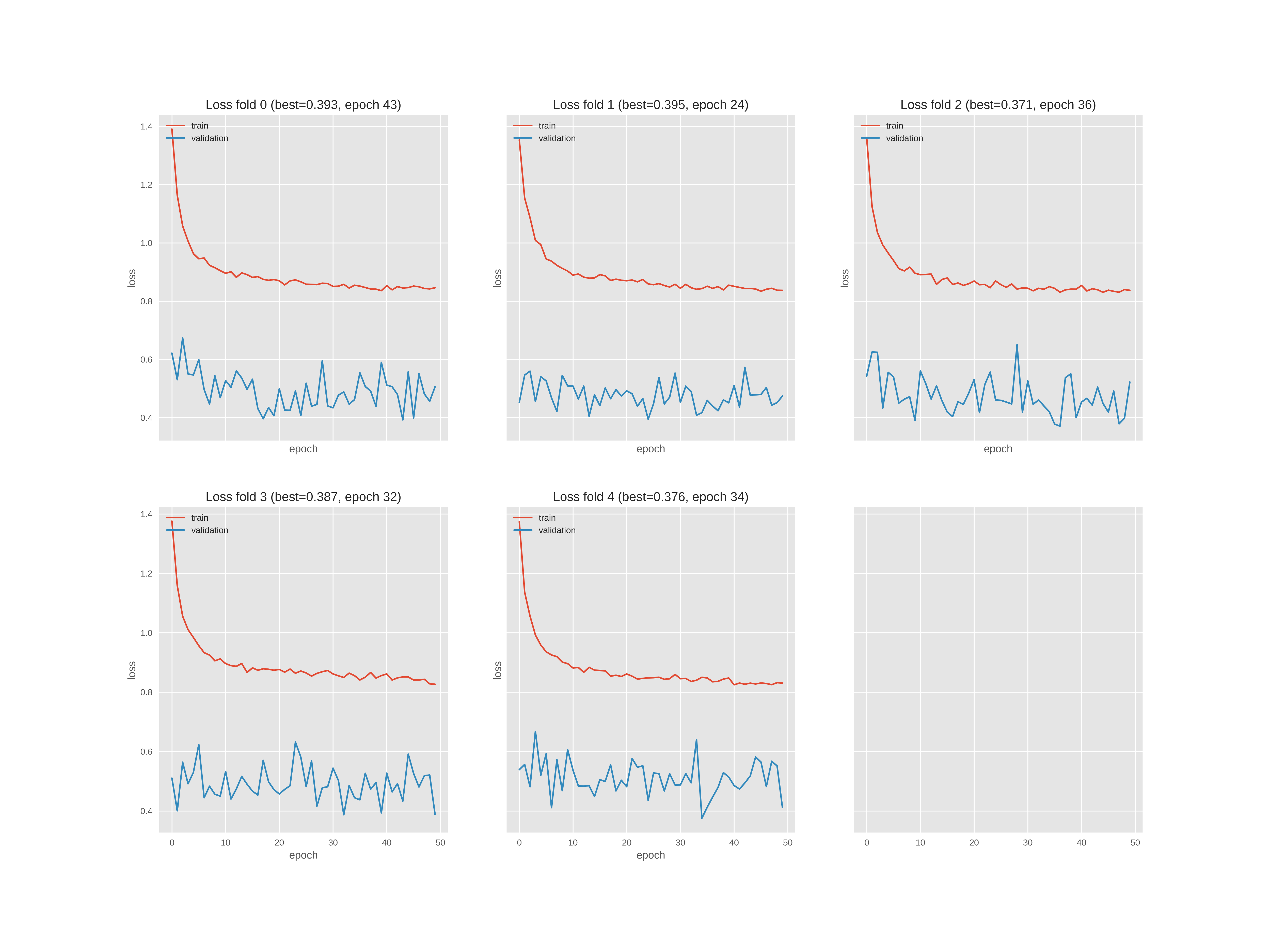}
    \caption{\label{loss_history} 5-Fold training loss history. }
\end{figure}
\subsection{Model performance}
Using this training configuration we obtained a cross validated Receiver Operating Characteristic Area Under the Curve (ROC AUC) of 0.8933 ($\pm 0.0032$) for the training set, and 0.8735 ($\pm 0.0025$) ROC AUC for the cross validation set. Using a $0.5$ decision threshold, the model reaches 75.423\% sensitivity at 82.776\% specificity.

\begin{figure}[h]
  
  \centering
    \includegraphics[width=\textwidth]{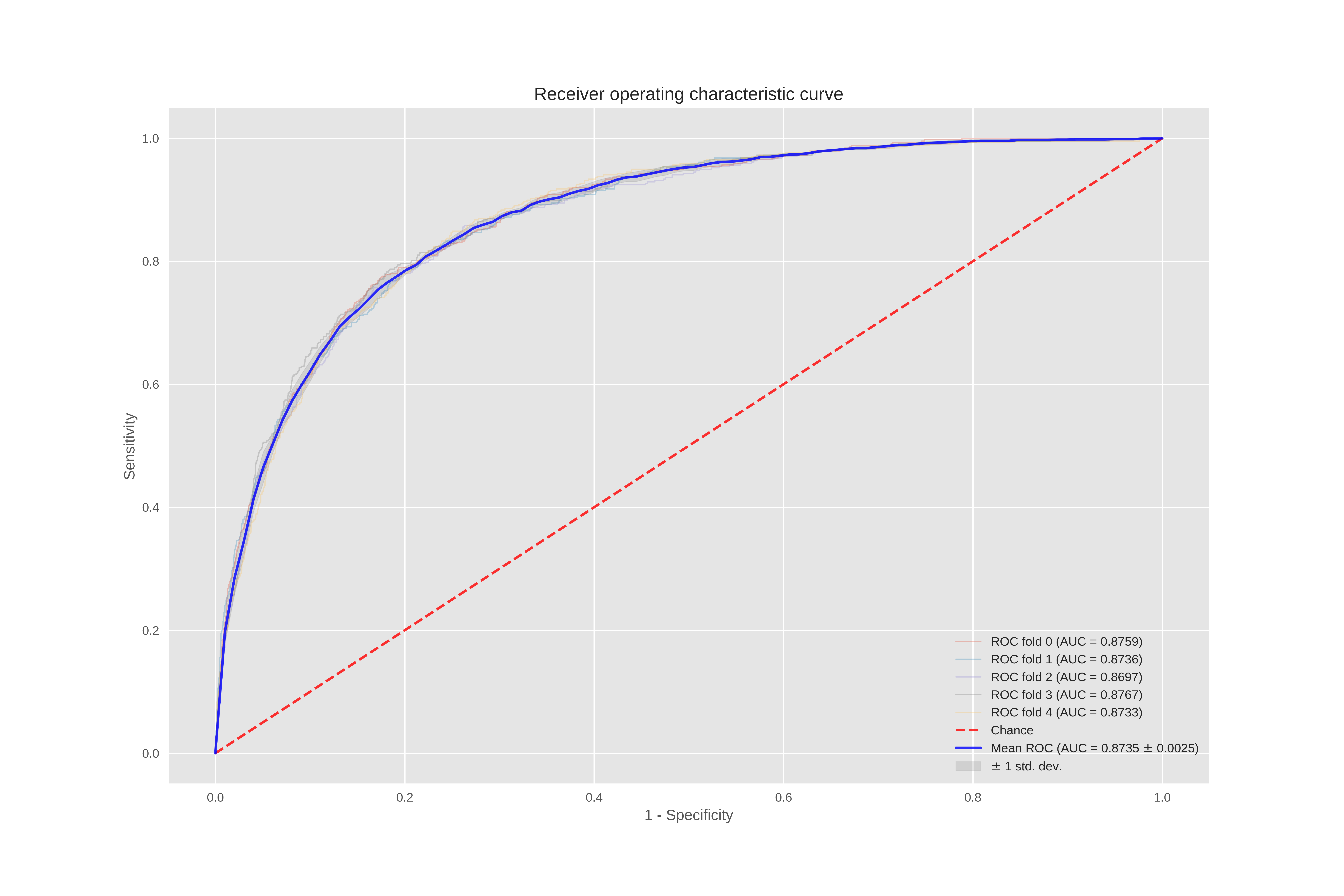}
    \caption{\label{validation_roc} ConvNet 5-Fold cross validated ROC AUC. }
\end{figure}

\subsection{Model interpretability}
We used the DeepLIFT implementation provided by its authors to compute our input feature importances from a model trained on one cross validation fold. We selected zero (empirical mean) as the reference value. We also computed importances for individual patients and at the dataset level, and created a series of visualizations to offer explanations for the predictions of the model. Visualizations are designed to combine patient features with their importance towards the predicted probability of death. Our visualizations constitute a form of \textit{post hoc interpretability} \cite{DBLP:journals/corr/Lipton16a} insofar as they try to convey how the model regards the inputs in terms of their impact on the predicted probability of death, without having to explain the internal mechanisms of our neural network, nor sacrificing predictive performance.

\begin{figure}[h]
  
  \centering
    \includegraphics[width=\textwidth]{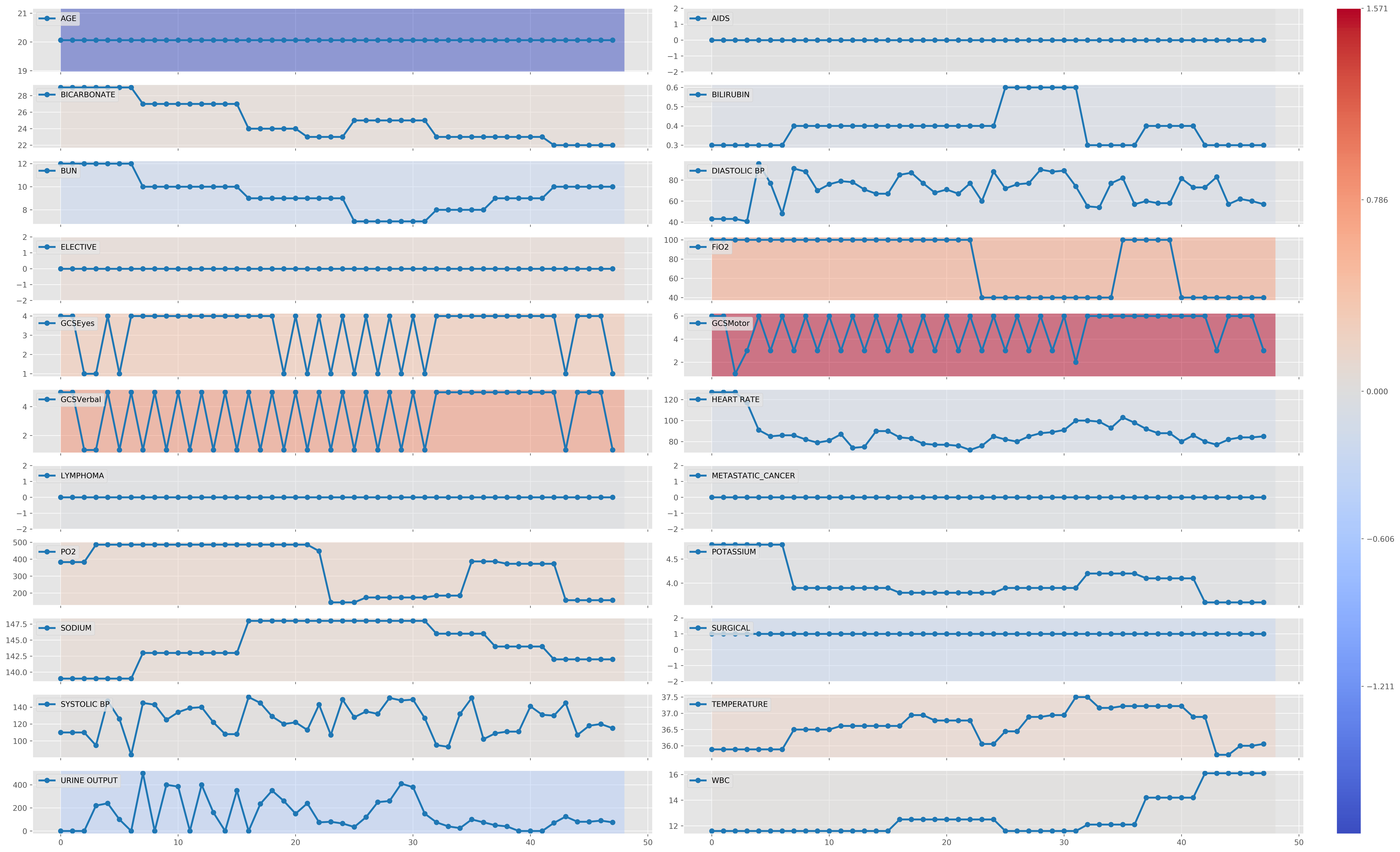}
    \caption{\label{explained_predictors} Marginal (total) predictor importance for a single patient. }
\end{figure}

\paragraph{Predictor importance}
Here we treated the patient tensor representation as an image and we grouped feature importance attribution semantically (i.e. observations belong to a particular predictor, as pixels on an image belong to an object) to find net contributions per predictor. Figure \ref{explained_predictors} shows the feature importances computed for a single patient (predicted probability of mortality: 0.5764, observed mortality: 1), summed over 48 hours for each individual predictor and then normalized over the predictor set. As mentioned this visualization shows the importance of each predictor as a whole, highlighting with red those predictors that contribute to a positive (death) prediction, and with blue those that contribute to a negative (survival) prediction. Since hourly importances can be either positive or negative in sign, it is possible that the total importance might be close to zero (gray background), even if the individual importances are not. We can clearly see that the network is assigning high positive importance to the components of the Glasgow Comma Scale - GCS, and high negative importance to the age of the patient. These are interesting because GCS values are shown to be abnormal, and the patient is very young (20 years old), and it is plausible that a young age is negatively correlated with mortality in the ICU. 

\begin{figure}[h]
  
  \centering
    \includegraphics[width=\textwidth]{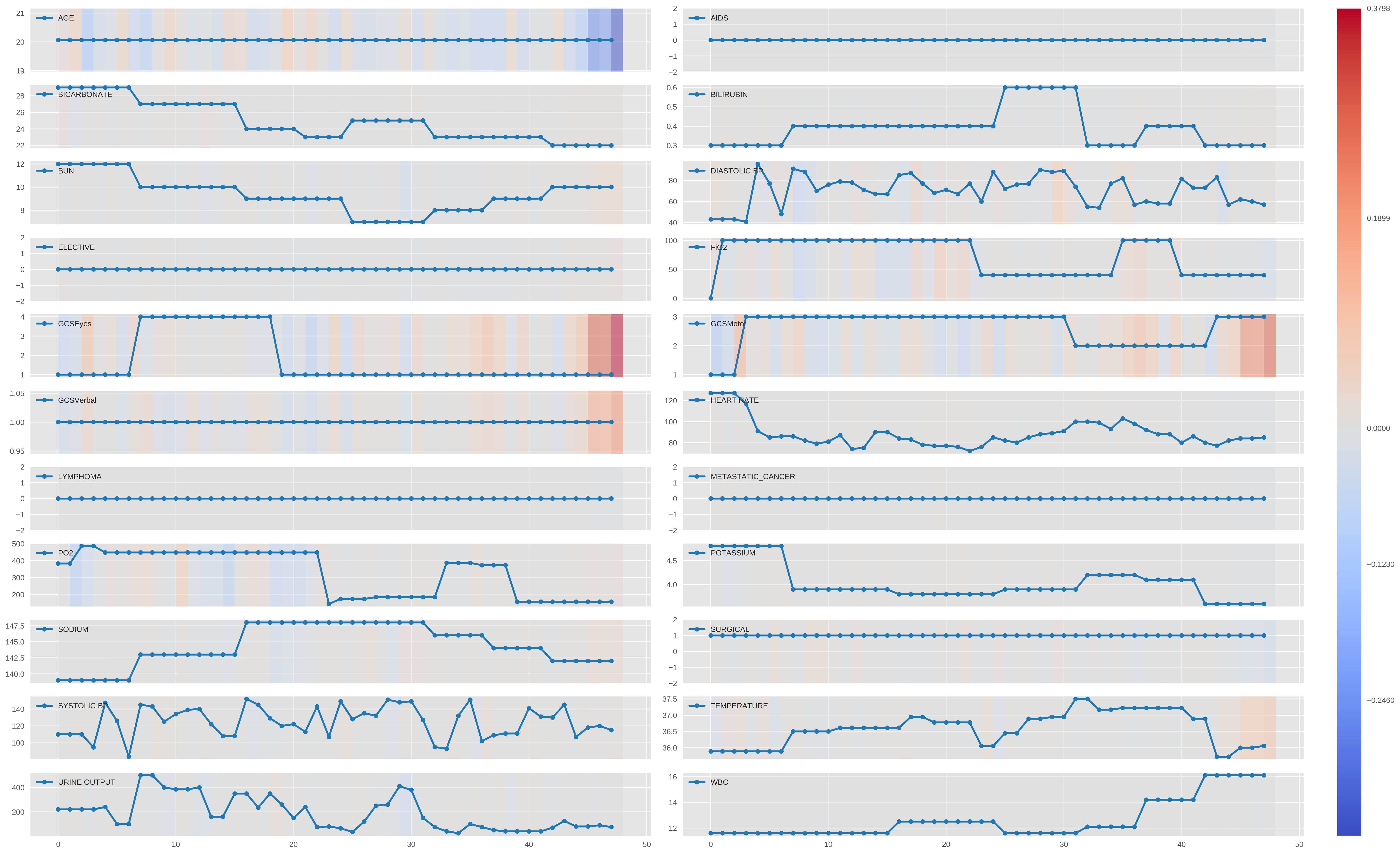}
    \caption{\label{explained_predictors_hourly} Predictor importance by hour for a single patient. }
\end{figure}

\paragraph{Predictor importance (hourly)}
In this visualization we further de-aggregate importance and show the individual approximate Shapley Values for each input value and hour (Figure \ref{explained_predictors_hourly}). We can see evidence for the non-linear dynamics the network has learned, as values from the same predictor have different importance across the temporal axis.

\begin{figure}[h]
  
  \centering
    \includegraphics[width=\textwidth]{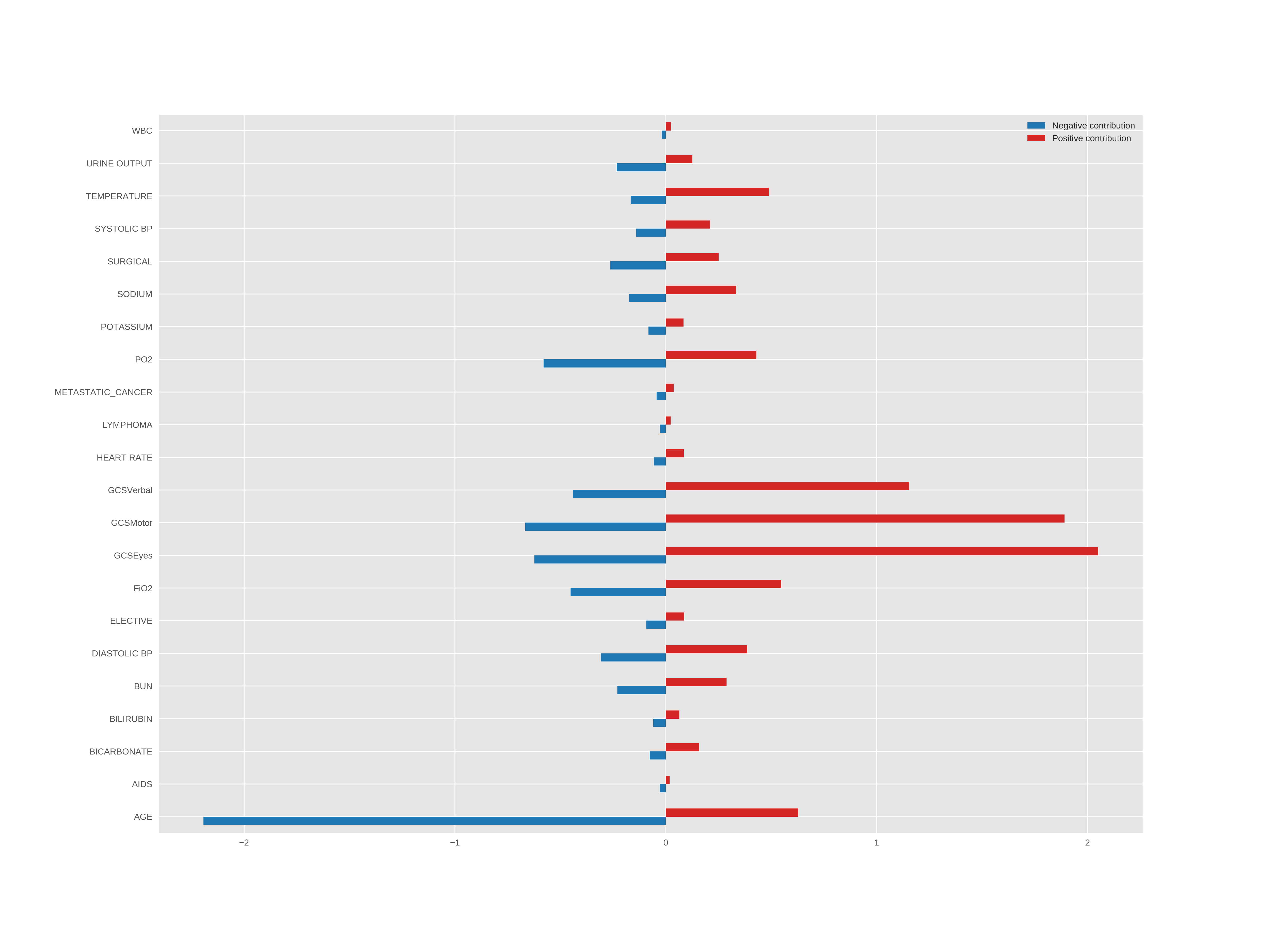}
    \caption{\label{explained_predictors_pos_neg} Negative and positive importance of each predictor for a single patient. }
\end{figure}

\paragraph{Positive and negative importance barplot} Alternatively we can treat up positive and negative importances separately to have a better sense of how each predictor affects the final prediction. Figure \ref{explained_predictors_pos_neg} shows a barplot with positive and negative importance grouped by predictor. 

\begin{figure}[h]
  
  \centering
    \includegraphics[width=\textwidth]{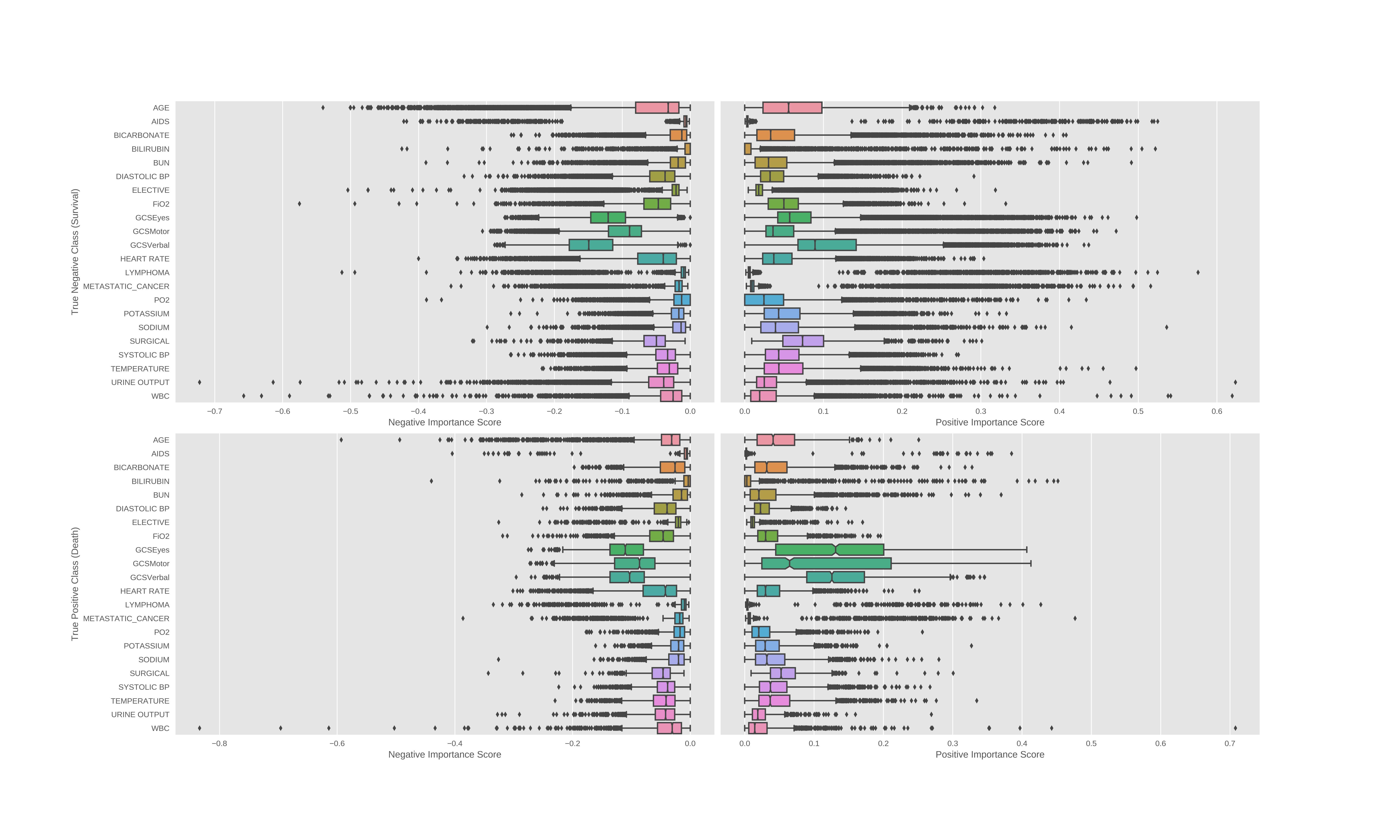}
    \caption{\label{dataset_explained_predictors_pos_neg} Boxplots for negative and positive predictor importance, grouped by true class at the dataset level.}
\end{figure}

\paragraph{Dataset-level feature importances} Additionally we computed importances for the validation set to offer interpretability at the dataset level. Figure \ref{dataset_explained_predictors_pos_neg} shows dataset-level statistics for the normalized positive and negative importance of each predictor.

\section{Discussion} Our ConvNet model shows strong performance on the MIMIC-III dataset with low variability across folds. Also performance over training and validation data are quite close, evidencing that our model exhibits signs of good generalization properties, as there is no serious overfitting ocurring (0.8933 ($\pm 0.0032$) vs 0.8735 ($\pm 0.0025$) ROC AUC). Validation performance reaches the state of the art for mortality prediction on MIMIC-III data and a comparable feature set (95\% CI [0.870396, 0.876604] against a 95\% CI [0.873706, 0.882894] corresponding to the results reported by \cite{PURUSHOTHAM2018}). Moreover our results show that a single convolutional architecture can handle both temporal and static features using simple time replication for the static inputs, instead of using a recurrent/feedforward hybrid architecture as in \cite{PURUSHOTHAM2018}.

\begin{table}[h]

\centering

\resizebox{\columnwidth}{!}{%
\begin{tabular}{ l l p{0.2\linewidth} l l l}
  \hline			
   Model & Type & Dataset & Task & ROC AUC & Interpretable? \\
   \hline	
   GRU-D \cite{DBLP:journals/corr/ChePCSL16} & Recurrent & MIMIC-III (99 features) & ICU mortality & 0.8527 $\pm $ 0.003 & No \\
   MMDL \cite{PURUSHOTHAM2018} & Hybrid & MIMIC-III (20 features) 	& ICU mortality & 0.8783 $\pm $ 0.0037 & No\\
   GBTmimic \cite{Ba:2014:DNR:2969033.2969123} & Hybrid + Gradient Boosted Trees & LA Children's Hospital PICU (48 features) &60-day mortality& 0.7898 $\pm$ 0.030	& Yes (dataset level) \\
   I-See-U & ConvNet & MIMIC-III (22 features) & ICU mortality & 0.8735 $\pm$ 0.0025 & Yes (patient and dataset level)\\
  \hline  
\end{tabular}%

}
\caption{\label{auccomparison} Our results and results reported by related works. ROC AUC results are mean and standard deviation from a 5-Fold cross validation run, except GBTmimic, which averages 5 different 5-Fold CV runs. }
\end{table}

Regarding model interpretability, the predictor marginal importance visualization shows that the model is attending to sensible features to predict mortality. As mentioned previously, the model is attending to the components of the GCS scale which show abnormal values and assigns them a positive contribution to mortality. PO2, FiO2, blood sodium and temperature are also regarded, to various degrees, as evidence favoring predicting mortality. On the other hand the patient age is regarded by the model as strong evidence against mortality, followed by urine output.

The marginal importance visualization allows us to see something interesting: the model assigns a negative net contribution to the fact that the patient was admitted after surgery, this is, the model regards the surgical admission as evidence for survival (however at the dataset level, median positive importance for surgical admissions are greater across classes than their negative counterpart, i.e. the model tends to see surgical admission as evidence for mortality). This could due to correlations present in the underlying dataset, or higher order interactions between predictors. The latter is attested by the predictor plus hour visualization, which shows that for static predictors, different observations accross time of the same predictor are assigned different contributions, sometimes with different sign. It is also worth noticing that while the patient's surgical admission is regarded as evidence for survival, the fact that the surgery was not an elective surgery is considered as evidence for mortality, which is sensible. However both input features must not be analysed separately (i.e. they correspond to a single concept in SAPS-II score \cite{Gall1993}). This is the kind of insight that interpretability efforts can reveal about black boxes, which is also absent in the majority of related works \cite{PURUSHOTHAM2018}.

Dataset-level analysis of feature importance show a high variance in attributed importance, both negative and positive. GCS  components tend to be the features with the most importance attributed (especially positive importance for patients that eventually died), followed by age. On the other hand, there are a number of features with both low positive and low negative mean importance. Presence of AIDS or lymphoma, are deemed by the ConvNet as not carrying much weight for predicting either survival or death. Also some of the other predictors have modest mean importances. This could signal a possibility to simplify the input feature set and get better predictive performance.

\section{Conclusions}

In this paper we presented ISeeU, a novel multi-scale convolutional network for interpretable mortality prediction inside the ICU, trained on MIMIC-III. We showed that our model offers state of the art performance, while offering visual explanations based on a concept from coalitional game theory, that show how important the inputs features are for the model's output. Such explanations are offered at the single patient level with different levels of de-aggregation, and at the dataset level, allowing for a more complete statistical understanding of how the model regards input predictors, compared to what related works have provided so far, and without resorting to auxiliary or surrogate models. We were able to show that a convolutional model can handle both temporal and static features at the same time without having to resort to hybrid neural architectures. We also showed that simple imputation techniques offer competitive performance without incurring in the computational costs associated with more complex approaches. 

\section{Acknowledgments}
GPU access and computing services provided by the Service and Cloud Computing Research Lab, Data Centre. Hosting services managed by Bumjun Kim, Senior Technician \href{mailto:bumjun.kim@aut.ac.nz}{bumjun.kim@aut.ac.nz} and ICT. The data centre is part of the School of Computer and Mathematical Sciences, Auckland University of Technology.

\section*{References}

\bibliography{phd}

\end{document}